\documentclass[preprint,12pt]{elsarticle}

\usepackage{amssymb}
\usepackage{lineno}
\usepackage{soul}
\usepackage{bm,amsthm,bbm,physics}
\newtheorem{theorem}{Theorem}
\newtheorem{corollary}{Corollary}
\newtheorem{lemma}{Lemma}
\newtheorem{definition}{Definition}
\newtheorem{remark}{Remark}
\usepackage{url}
\usepackage{float}
\usepackage{hyperref}
\hypersetup{hidelinks,pdfauthor=whatever}
\usepackage{xcolor}
\usepackage[hypcap=false]{caption}
\journal{}
\begin{document}

\begin{frontmatter}



\title{‌One-shot Generative Distribution Matching for Augmented RF-based UAV Identification}


% Authors with multiple affiliations
\author[1,2]{Amir Kazemi}
\author[3]{Salar Basiri}
\author[1,4,5]{\\Volodymyr Kindratenko}
\author[3]{Srinivasa Salapaka\corref{cor1}}

% Affiliations
\affiliation[1]{organization={National Center for Supercomputing Applications}}
\affiliation[2]{organization={Department of Civil and Environmental Engineering}}
\affiliation[3]{organization={Department of Mechanical Science and Engineering}}
\affiliation[4]{organization={Department of Computer Science}}
\affiliation[5]{organization={Department of Electrical and Computer Engineering},
            addressline={\\University of Illinois at Urbana-Champaign}, 
            city={Urbana},
            postcode={61801}, 
            state={IL},
            country={USA}}

% Corresponding authors
\cortext[cor1]{Correspondence to: \{kazemi2, sbasiri2, kindrtnk, salapaka\}@illinois.edu}

%% Abstract
\begin{abstract}
This work addresses the challenge of identifying Unmanned Aerial Vehicles (UAV) using radiofrequency (RF) fingerprinting in limited RF environments. The complexity and variability of RF signals, influenced by environmental interference and hardware imperfections, often render traditional RF-based identification methods ineffective. To address these complications, the study introduces the rigorous use of one-shot generative methods for augmenting transformed RF signals, offering a significant improvement in UAV identification. This approach shows promise in low-data regimes, outperforming deep generative methods like conditional generative adversarial networks (GANs) and variational auto-encoders (VAEs). The paper provides a theoretical guarantee for the effectiveness of one-shot generative models in augmenting limited data, setting a precedent for their application in limited RF environments. This research contributes to learning techniques in low-data regime scenarios, which may include atypical complex sequences beyond images and videos. The code and links to datasets used in this study are available at \url{https://github.com/amir-kazemi/uav-rf-id}.
\end{abstract}


%\item Theoretical upper bound on distributional distance between synthetic and real data is derived and validated, ensuring fidelity of generated samples for reliable UAV classification across various scenarios.
%\item Distribution matching by a one-shot generative model (GPDM) outperforms deep learning methods (CGAN, CVAE) in data-scarce UAV identification, achieving up to 25\% higher accuracy when tested on 10\% of training data in limited RF environments.

%% Keywords

\end{frontmatter}


%\begin{linenumbers}
\section{Introduction}
The growing integration of Unmanned Aerial Vehicles (UAVs), commonly known as drones, into various sectors, ranging from surveillance to logistics, has underscored the critical need for their cybersecurity. As such, employing radiofrequency (RF) fingerprinting becomes crucial for identifying legitimate versus malicious devices, particularly in environments with constraints \cite{yaacoub2020security,khan2022detection}.
The complexity and variability of RF signals, influenced by factors such as temperature, humidity \cite{lim2020review}, environmental obstructions \cite{sinha2020rss,zhang2023rf}, and hardware imperfections \cite{gul2022fine}, pose significant challenges for consistent UAV identification and tracking. These conditions can interfere with RF signal clarity and strength, leading to a limited RF environment where the number of usable signals is reduced. Traditional RF-based identification methods, although effective in open settings, often struggle in such limited RF environments—marked by dense foliage, urban canyons, adverse weather, or potentially exacerbated by hardware malfunctions—making reliable UAV detection and classification particularly challenging.

The advent of machine learning techniques has opened new avenues for enhancing UAV identification in these challenging RF environments. By leveraging deep learning algorithms, it is possible to extract and analyze complex patterns within the RF signals that are typically obscured or distorted due to environmental factors \cite{al2019rf,nemer2021rf,nie2021uav,wang2022rf}.
However, the application of data-intensive deep learning methods does not inherently resolve the issue of data scarcity in compromised environments. This necessitates the development of reliable techniques for augmenting RF signals or their transformations in situations with limited data. The challenge arises mainly because current data augmentation techniques are predominantly tailored for computer vision applications, and their extension to other types of sequences may not yield the anticipated outcomes due to the complex semantics of the signals.

In this study, we rigorously employ a one-shot generative method, inspired by computer vision techniques, for the augmentation of DFT-transformed RF signals across various training data sizes which simulates low-data regimes posed by limited RF environments. The results demonstrate a significant improvement in evaluation metrics for the UAV identification, outperforming other frameworks such as conditional generative adversarial networks (GAN) and variational auto-encoders (VAE) in addressing low-data regime circumstances.

The work is therefore structured as follows: First, we investigate the ability of one-shot generative models for augmenting a dataset of sequences. Particularly, we provide the mathematical definition of sequence, subsequence, one-shot generative models, and distance metrics. We establish an upper bound for the Wasserstein distance between the real and generated sequences, considering the expected distances of their subsequences. We also validate our general guarantee by intuitive special cases of sequences and subsequences.
Secondly, we go through the UAV identification experiment and elaborate on the selected dataset, the one-shot generative model, benchmark conditional deep generators, the classification pipeline, and the evaluation method. Finally, we discuss the results which include the accuracy, precision, recall, and F1-score of the classification task for all designed cases and methods.

\section{Related generative techniques}

Synthetic data generation is a crucial field of study in machine learning and data science, with potential applications spanning a diverse array of disciplines. In many cases, obtaining large amounts of high-quality, labeled data is a significant challenge, and this scarcity can lead to biased or skewed results when training models \cite{nikolenko2021synthetic}. Data generation methods can help create edge cases, rare events, and other scenarios that are difficult to collect in real-world circumstances, allowing for more robust model training \cite{libes2017issues}. Distribution matching, an integral part of synthetic data generation, involves adjusting the synthetic data's statistical properties to closely resemble those of the real data, thereby enhancing the validity and effectiveness of the models trained on this data \cite{yuan2023real}.

The past decade marked a dramatic increase in efforts to generate sequences for a broad range of data, from images and videos to audio, text, and spatiotemporal physical states. This surge in sequence generation can be attributed largely to the recent advances in convolutional and recurrent neural networks (CNN and RNN) \cite{rawat2017deep,gu2018recent,yu2019review,graves2013generating} and deep generative frameworks such as GAN \cite{goodfellow2020generative} and VAE \cite{kingma2013auto}. At the heart of these architectures and frameworks lie deep neural networks (DNN). However, when training generalizable DNNs using Empirical Risk Minimization (ERM) \cite{zhang2017mixup}, it is paradoxical to depend on deep generative models that require large amounts of data to address data scarcity in subsequent tasks. This situation highlights the necessity to explore alternative approaches that can address the challenges of data-intensive models, utilizing a concept known as Vicinal Risk Minimization (VRM).

VRM is a training principle that improves generalization in machine learning models by considering the vicinity of the training samples, hypothesizing that slight perturbations around real data points can yield more robust and better-performing models \cite{chapelle2000vicinal}. This concept seamlessly leads to the notion of data augmentation, a widely-adopted strategy in deep learning, especially in fields like computer vision where common methods of data augmentation include random cropping, rotation, and brightness adjustments \cite{shorten2019survey}.

With the emergence of one-shot generative models, a fresh motivation surfaces; harnessing their potential to generate realistic variations of training samples might offer a more sophisticated approach to VRM. Leveraging such generative capacities could potentially revolutionize the landscape of data augmentation and VRM practices. Recent one-shot generative models \cite{shaham2019singan,shocher2019ingan,hinz2021improved,asano2019critical}, and their non-deep successors \cite{granot2022drop,elnekave2022generating,haim2022diverse}, generate patch-similar images and videos from a single target.
Considering the patch-similarity as a vicinity criterion, one may therefore explore the ability of one-shot generative models for augmenting an entire dataset inspired by the VRM principle.

In this study, our focus is on the application of one-shot generative models to sequences, which serve as a universal structure for ordered data, extending from images to time series. We aim to establish theoretical limits on their ability to augment datasets. Our work utilizes the Generating by Patch Distribution Matching (GPDM) algorithm, detailed in \cite{elnekave2022generating}. GPDM is designed to minimize the Wasserstein distance between patches of both real and generated images. Furthermore, this algorithm employs a multi-scale hierarchical strategy, facilitating an early understanding of the global image structure during the initial training stages. GPDM has emerged as the most efficient one-shot generative model capable of generating diverse samples, primarily due to its optimization based on Sliced Wasserstein distance (SWD). In contrast, other one-shot generative models either lack speed or do not employ metric distances, potentially compromising sample diversity \cite{elnekave2022generating}. For instance, Generative Patch Nearest Neighbors (GPNN) ~\cite{granot2022drop} utilizes K-Nearest Neighbors (KNN), a similarity measure that is non-metric and results in less diverse samples compared to GPDM. Additionally, deep models like SinGAN \cite{shaham2019singan} require extensive training times for each sample, which is inefficient compared to the model-free approach of GPDM. 

\section{Sequence and subsequence similarity} \label{similarities}

\begin{figure}[ht]
    \centering
    \includegraphics[width=\columnwidth]{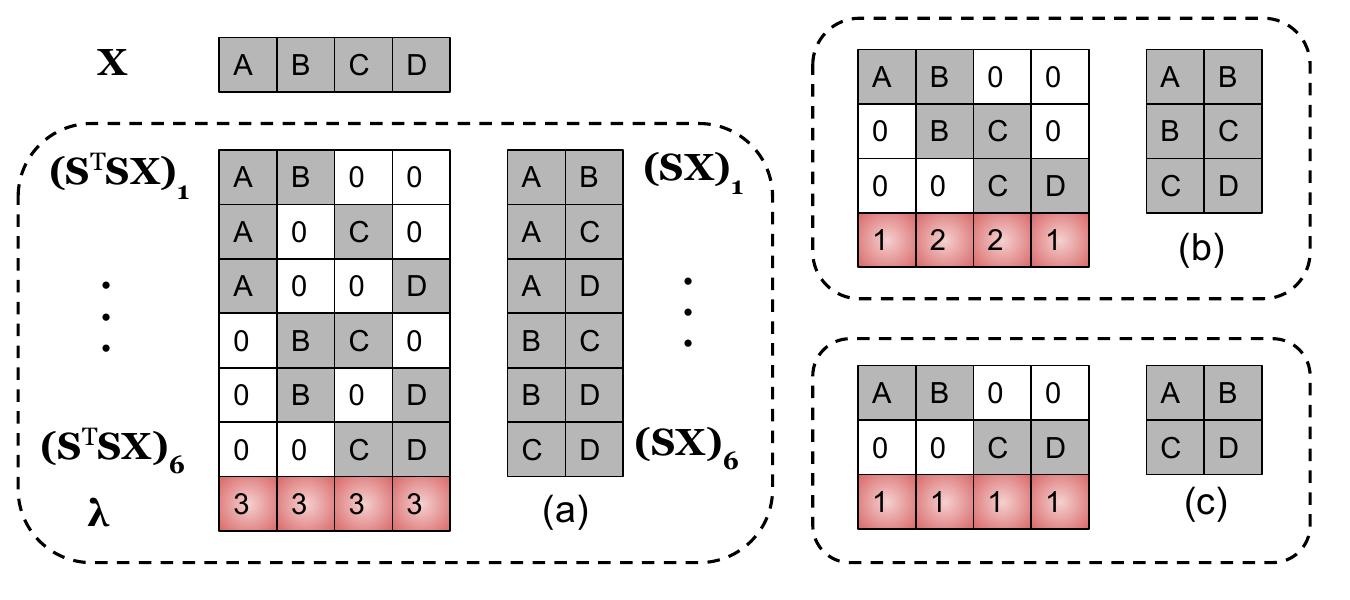}
    \caption{Sequence $\mathbf{X}$, subsequence $\mathbf{SX}$, and projection $\mathbf{S}^T\mathbf{SX}$ for $d=4, d'=2$: (a) $\tilde\Omega_S=\Omega_S$ is the set of all subsequences of length $d'=2$ which includes six subsequences, (b) and (c) $\tilde\Omega_S\subset \Omega_S$ is the set of substrings of length $d'=2$.}
    \label{fig:data_structure}
\end{figure}

This section outlines the derivation of guarantees for creating synthetic datasets through one-shot generation. It confirms that these datasets, when generated using one-shot methods, maintain a distribution close to the original dataset within a set threshold. These methods focus on maximizing the similarity between patches of the target and generated images, videos, and other sequences. The concept of a sequence is therefore crucial for thoroughly understanding this work, especially when dealing with RF signals or their Fourier transformations. Such data are interpreted as sequences, where their patches are akin to substrings, or more broadly, subsequences. Given this importance, our focus primarily lies on defining sequences, subsequences, and projections, as depicted in Fig. \ref{fig:data_structure}. To achieve this, we initially define selection matrices, as subsequences and projections are derived from sequences using these matrices. Key variables and terms throughout this paper are listed in the nomenclature Table \ref{tab:nomenclature} for easy reference.

\begin{table}
\caption{Nomenclature}
\label{tab:nomenclature}
\setlength{\tabcolsep}{3pt}
\begin{tabular}{|l|l|}
\hline
Symbol & Description \\
\hline
$d$ & Length of sequence $\mathbf{X}$ \\
$\mathbf{X} ,\mathbf{X'}$ & IID random sequences defined as vectors on $(\Omega_X\subseteq \mathbb{R}^d, P_X)$ \\
$d'$ & Length of subsequences, where $d'\leq d$ \\
$\mathbf{I}_d$ & $d\times d$ identity matrix \\
$\mathbf{S}, \mathbf{S}'$ & IID $d'\times d$ random selection matrices defined on $(\Omega_S, P_S)$, see Def. \ref{Ldef}. \\
$\mathbf{s}$ & A realization of $\mathbf{S}$ \\
$\mathbf{SX}$ & A $d'$-dimensional vector representing a subsequence of $\mathbf{X}$ \\
$\mathbf{S}^T\mathbf{SX}$ & A $d$-dimensional vector representing a projection of $\mathbf{X}$ \\
$\tilde\Omega_S$ & Any subset of $\Omega_S$ satisfying Def. \ref{Ldef}. \\
$\mathbf{Z}$ & Noise defined on $(\Omega_Z \subseteq\mathbb{R}^\zeta, P_Z)$ \\
$G$ & One-shot generative model, see Def. \ref{Gdef} \\
$P_G$ & The pushforward measure of $G (\mathbf X, \mathbf Z)$ \\
\hline
\multicolumn{2}{|p{\dimexpr\linewidth-2\tabcolsep}|}{To streamline notation, we omit the event space in a probability space. The power set serves as the event space for countable sample spaces, while the Borel 
$\sigma$-algebra is assumed for uncountable ones.}\\
\hline
\end{tabular}
\end{table}

\begin{definition}\label{Ldef}
We define \textbf{S} as a selection matrix on the probability space \((\Omega_S, P_S)\), where \(E\) is an event composed of such selections, i.e. $E \subseteq \Omega_S$. Specifically, we have
\begin{equation}\label{samples_S}
\begin{split}
\Omega_S & = \{ \mathbf{I_d}(J,:) | 1 \leq j_1 < j_2 < \ldots < j_{d'} \leq d\}, \\ P_S(E) & =|E||\Omega_S|^{-1}.
\end{split}
\end{equation}
We may generally assume a subset of $\Omega_S$ as the sample space to have $(\tilde\Omega_S \subseteq \Omega_S, P_S)$ equipped with $P_S(E) = |E||\tilde\Omega_S|^{-1}$ and $E \subseteq \tilde\Omega_S$, provided that $\tilde\Omega_S$ is sufficiently large to fully recover a sequence from its projections, i.e.
\begin{equation}\label{events_S}
\mathbf X 
= \textstyle
\sum_{\mathbf s \in \tilde\Omega_S} \mathbf s ^T \mathbf s \mathbf X \oslash \bm\lambda,
\end{equation}
where $\oslash$ denotes Hadamard division and
\begin{equation}
\bm\lambda=\textstyle\sum_{\mathbf s \in \tilde\Omega_S} \mathbf s ^T \bm 1_{d'\times 1}.
\end{equation}
\end{definition}
\medskip
According to the definition, $\Omega_S$ is the set of all matrices created by choosing $d'$ rows from the identity matrix $\mathbf{I}_d$. The probability distribution for singleton events in $\tilde\Omega_S \subseteq \Omega_S$ is uniform. Furthermore, the subset $\tilde\Omega_S$ must be sufficiently large to allow for the recovery of the sequence $\mathbf{X}$ from the projections $(\mathbf{s}^T \mathbf{s} \mathbf{X})$, as per  (\ref{events_S}). This suggests that $\bm\lambda \in \mathbb{N}^d$, since $\bm\lambda$ represents the frequency of occurrences of elements from $\mathbf{X}$ in these projections, as illustrated in Fig. \ref{fig:data_structure}. Based on this definition, we can proceed to define the behavior of a one-shot generative model. This model is designed to produce subsequence-similar sequences for every target sequence within the dataset. Before proceeding, we elucidate our understanding of similarity by referencing a conventional metric distance used to compare probability distributions.

The Wasserstein distance is a powerful metric for quantifying the difference between two probability distributions. This metric, often referred to as the \textit{earth mover's distance}, provides an intuitive measure of the effort required to transform one distribution into another by considering the cost of moving distribution mass. Unlike other distance metrics that measure differences at fixed points, the Wasserstein distance considers the geometry of the space by accounting for the movement of distribution mass across it. To facilitate a deeper understanding, we will break down the mathematical formulation of the Wasserstein distance.
\\
\begin{definition}[Distance Metric for Distributions]
The Wasserstein distance between sequences is defined as:
\begin{equation}
 W(P_X, P_Y) =\inf_{\gamma \in \Gamma}\mathbb{E}_{(X,Y)\sim \gamma} \norm{\mathbf X-\mathbf Y}
\end{equation}
where $\norm{\cdot}$ denotes the L1 norm throughout this work, and $\Gamma$ is the set of couplings (joint distributions) whose marginals are $P_X$ and $P_Y$. For notational brevity, $\inf_{\gamma \in \Gamma}\mathbb{E}_{(X,Y)\sim \gamma}$ is denoted as $\inf\mathbb{E}_{X,Y}$ through this work.
\end{definition}
\medskip

We incorporate the Wasserstein metric to establish an upper bound on the distance between the probability distributions of real and synthetic data later. However, it is essential to first define a key feature of the one-shot generative model, denoted as \( G(\mathbf{X}, \mathbf{Z}) \). This function is typically designed to generate sequences influenced by the target \( \mathbf{X} \) and modulated by the noise \( \mathbf{Z} \). We impose a constraint on \( G(\mathbf{X}, \mathbf{Z}) \): it is the limit on the Wasserstein distance between the subsequences of \( G(\mathbf{X}, \mathbf{Z}) \) and \( \mathbf{X} \), averaged over both \( \mathbf{X} \) and \( \mathbf{Z} \). This constraint ensures that the sequences synthesized by the model maintain a minimum level of similarity to the target sequence \( \mathbf{X} \), in terms of their internal distributional properties. This characteristic is critical for preserving the integrity of the generative process, ensuring that while the model introduces variability through \( \mathbf{Z} \), it also adheres to the internal information of \( \mathbf{X} \).
\\
\begin{definition}\label{Gdef}[One-shot Generative Model]
$G (\mathbf X, \mathbf Z)$ is a generated sequence with $G:\mathbb{R}^d \times \mathbb{R}^\zeta \rightarrow \mathbb{R}^d$ as the one-shot generative function. We assume $\forall \delta>0$, there exists a measurable function $G$ that admits the following property:
\begin{equation}
\label{def1}\mathbb{E}_X\mathbb{E}_Z \inf \mathbb{E}_{S,S'}\norm{\mathbf{S}\mathbf X-\mathbf{S}'G (\mathbf X, \mathbf Z)}\leq \delta.
\end{equation}
\end{definition}

\begin{figure}[ht]
    \centering
    \includegraphics[width=\columnwidth]{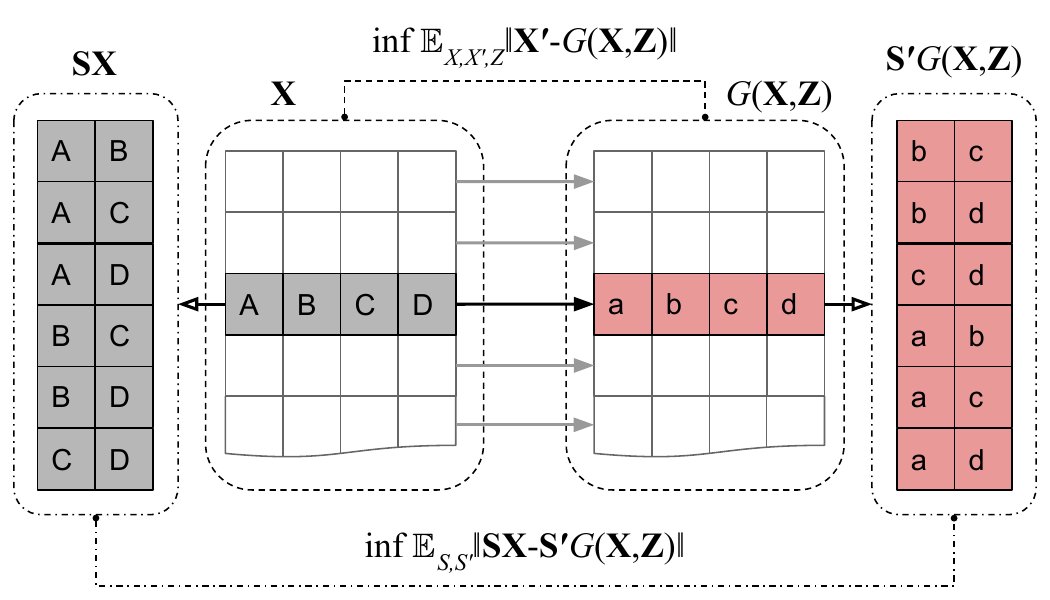}
    \caption{
    A one-shot generative model takes in real sequences $\mathbf X$ and outputs synthetic sequences $G(\mathbf X, \mathbf Z)$. The distance metric employed is optimal transport, specifically the Wasserstein distance, applied between the distributions of sequences and subsequences, as detailed in the accompanying formulation.}
    \label{fig:vrm}
\end{figure}

Using the previously defined concepts, we can outline the broader scope of our approach as depicted in Fig. \ref{fig:vrm}. This schematic illustrates the generation of synthetic data using the one-shot generative model and represents the distance metric between sequences and subsequences. In the following lemma, we establish a bound for the expected distance between the projections of two sequences: $\mathbf{X}'$ and $G(\mathbf{X}, \mathbf{Z})$. It is important to note that $G(\mathbf{X}, \mathbf{Z})$ is generated from $\mathbf{X}$, which is IID to $\mathbf{X}'$. The relevance of this lemma might not be immediately apparent. However, its significance becomes evident as it facilitates the proof of the ensuing theorem, which constitutes the core of this section.
\\

\begin{lemma}
Let $G$ be a measurable function. Then, for all $\mathbf X, \mathbf X'$ on $(\Omega_X, P_X)$, $\mathbf Z$ on $(\Omega_Z, P_Z)$, and $\mathbf S, \mathbf S'$ on $(\tilde\Omega_S, P_S)$ we have:
\begin{equation}
\mathbb{E}_S\norm{\mathbf{S}^T\mathbf{S}\mathbf{X'}-\mathbf{S}^T\mathbf{S}G (\mathbf X, \mathbf Z)} \leq 
\inf \mathbb{E}_{S,S'}\norm{\mathbf{S}\mathbf{X}-\mathbf{S}'G (\mathbf X, \mathbf Z)}
\end{equation}
\end{lemma}
\begin{proof}
On one hand, the distance between projections induced by the same selection matrix \( (\mathbf S')\) equals to the distance between corresponding subsequences, i.e.
\begin{equation} \label{lem:1}
\norm{\mathbf{S}'^T\mathbf{S}'\mathbf{X'}-\mathbf{S}'^T\mathbf{S}'G (\mathbf X, \mathbf Z)}
=
\norm{\mathbf{S}'\mathbf{X'}-\mathbf{S}'G (\mathbf X, \mathbf Z)} .
\end{equation}
On the other hand, the triangle inequality gives:
\begin{equation} \label{lem:2}
\norm{\mathbf{S}'\mathbf{X'}-\mathbf{S}'G (\mathbf X, \mathbf Z)} \leq 
\norm{\mathbf{S}\mathbf{X}-\mathbf{S}'G (\mathbf X, \mathbf Z)}  
+
\norm{\mathbf{S}\mathbf{X}-\mathbf{S}'\mathbf{X'}}.
\end{equation}
Therefore, Eqs. (\ref{lem:1}) and (\ref{lem:2}) give:
\begin{equation}
\begin{split}
& \norm{\mathbf{S}'^T\mathbf{S}'\mathbf{X'}-\mathbf{S}'^T\mathbf{S}'G (\mathbf X, \mathbf Z)} \\ &
\leq 
\norm{\mathbf{S}\mathbf{X}-\mathbf{S}'G (\mathbf X, \mathbf Z)}  
+
\norm{\mathbf{S}\mathbf{X}-\mathbf{S}'\mathbf{X'}}, 
\end{split}
\end{equation}
which for every coupling $(\mathbf S, \mathbf S')$ becomes:
\begin{equation}
\begin{split}
& \mathbb{E}_{S'}\norm{\mathbf{S}'^T\mathbf{S}'\mathbf{X'}-\mathbf{S}'^T\mathbf{S}'G (\mathbf X, \mathbf Z)}
 \\ & \leq 
\mathbb{E}_{S,S'}\norm{\mathbf{S}\mathbf{X}-\mathbf{S}'G (\mathbf X, \mathbf Z)}  
+
\mathbb{E}_{S,S'}\norm{\mathbf{S}\mathbf{X}-\mathbf{S}'\mathbf{X'}}.
\end{split}
\end{equation}
Minimizing both sides for the coupling $(\mathbf S, \mathbf S')$, we have:
\begin{equation}
\inf \mathbb{E}_{S,S'}\norm{\mathbf{S}\mathbf{X}-\mathbf{S}'\mathbf{X'}} = 0
\end{equation}
as a result of choosing a coupling satisfying $\mathbf X=\mathbf X'$ and $\mathbf S = \mathbf S'$. Therefore, with an abuse of notation on LHS (using $\mathbf S$ instead of $\mathbf S'$), we prove the lemma.
\end{proof}

\begin{theorem}
Assume that $\forall \delta>0$, there exists a measurable function $G$ such that: $\mathbb{E}_X\mathbb{E}_Z\inf \mathbb{E}_{S,S'}\norm{\mathbf{S}\mathbf{X}-\mathbf{S}'G (\mathbf X, \mathbf Z)}\leq \delta$. Then we have
\begin{equation}\label{general_bound}
W(P_X,P_G) \leq  \textstyle
|\tilde\Omega_S|\bm\lambda_{\min}^{-1}\delta.
\end{equation}
\end{theorem}

\begin{proof}
By definition in Eq \eqref{events_S}, we have
\begin{equation}
\begin{split}
& \Vert \mathbf X' - G (\mathbf X, \mathbf Z)\Vert \\ = &\norm{ \textstyle \sum_{\mathbf s \in \tilde\Omega_S} (\mathbf s^T\mathbf{sX'} - \mathbf s^T\mathbf{s}G (\mathbf X, \mathbf Z)) \oslash
\bm\lambda }
\end{split}
\end{equation}
which, considering the fact that $\bm\lambda \in \mathbb{N}^d$, it yields
\begin{equation}
\begin{split}
& \Vert \mathbf X' - G (\mathbf X, \mathbf Z)\Vert
 \\ \leq &
\textstyle \bm\lambda_{\min}^{-1} \sum_{\mathbf s \in \tilde\Omega_S} \norm{\mathbf s^T\mathbf{sX'} - \mathbf s^T\mathbf{s}G (\mathbf X, \mathbf Z)
}\\
= &
\textstyle |\tilde\Omega_S|\bm\lambda_{\min}^{-1}
\mathbb{E}_S \norm{\mathbf S^T\mathbf{SX'} - \mathbf S^T\mathbf{S}G (\mathbf X, \mathbf Z)
}
\end{split}
\end{equation}
Using Lemma 1, we get
\begin{equation}
\Vert \mathbf X' - G (\mathbf X, \mathbf Z)\Vert
\leq  \textstyle |\tilde\Omega_S|\bm\lambda_{\min}^{-1}
\inf \mathbb{E}_{S,S'}\norm{\mathbf{S}\mathbf{X}-\mathbf{S}'G (\mathbf X, \mathbf Z)}.
\end{equation} 
When we minimize the expected value over the coupling $(\mathbf X, \mathbf X', \mathbf Z)$ for both sides, we have
\begin{equation} \label{theorem:eq}
\inf \mathbb{E}_{X, X', Z} \Vert \mathbf X' - G (\mathbf X, \mathbf Z)\Vert \leq  |\tilde\Omega_S|\bm\lambda_{\min}^{-1}\delta,
\end{equation}
because
\begin{equation}
\begin{split}
\inf \mathbb{E}_{X, Z} \inf \mathbb{E}_{S,S'}\norm{\mathbf{S}\mathbf{X}-\mathbf{S}'G (\mathbf X, \mathbf Z)} \\
 \leq \mathbb{E}_X\mathbb{E}_Z \inf \mathbb{E}_{S,S'}\norm{\mathbf{S}\mathbf{X}-\mathbf{S}'G (\mathbf X, \mathbf Z)}
\stackrel{\mathrm{Def.} \; (\ref{Gdef})}{\leq} \delta.
\end{split}
\end{equation}
The LHS of  (\ref{theorem:eq}) defines the Wasserstein distance, thereby proving the theorem. To satisfy the theorem, the optimal coupling over $(\mathbf X, \mathbf X', \mathbf Z)$ must adhere to the marginal distributions $P_X$ for $\mathbf X$ and $\mathbf X'$, $P_Z$ for $\mathbf Z$, and the pushforward measure $P_{G}$ for $G(\mathbf X, \mathbf Z)$.
\end{proof}
We may gain more insight from the bound in  (\ref{general_bound}) if we consider special cases of $\tilde\Omega_S$.
In the next corollary, we will see the behavior of the bound for $\tilde\Omega_S=\Omega_S$ and, in the following corollary, we consider the two-dimensional case of sequences. The latter is critical for using existing generative models, as most of them are designed for two-dimensional arrays (such as images).
\\

\begin{corollary}\label{cor1}
Assume that $\forall \delta>0$, there exists a measurable function $G$ such that: $\mathbb{E}_X\mathbb{E}_Z\inf \mathbb{E}_{S,S'}\norm{\mathbf{S}\mathbf{X}-\mathbf{S}'G (\mathbf X, \mathbf Z)}\leq \delta$. Then, if $\tilde\Omega_S=\Omega_S$ we have
\begin{equation}
W(P_X,P_G) \leq
\left( d\mathbin{/}d'\right)\delta.
\end{equation}
\end{corollary}

\begin{proof}
If $\tilde\Omega_S$ equals the set of all subsequences of length $d'$, i.e. $\Omega_S$, then we have
\begin{equation}
|\tilde\Omega_S| = \binom{d}{d'} \; , \; \bm\lambda = \binom{d-1}{d'-1} \mathbf{1}^{d\times 1}.
\end{equation}
Therefore, Theorem 1 gives
\begin{equation}
W(P_X,P_G) \leq
\left( \binom{d}{d'}\mathbin{/}\binom{d-1}{d'-1} \right)\delta = \left( d\mathbin{/}d'\right)\delta
\end{equation}
which proves the corollary. The result is intuitive, because as the length of subsequences $d'$ tends to the length of the sequence $d$, the bound on the distributional distance of sequences and subsequences matches each other, i.e. $\delta$.
\end{proof}

\begin{corollary}\label{cor2}
Assume that $\forall \delta>0$, there exists a measurable function $G$ such that: $\mathbb{E}_X\mathbb{E}_Z\inf \mathbb{E}_{S,S'}\norm{\mathbf{S}\mathbf{X}-\mathbf{S}'G (\mathbf X, \mathbf Z)}\leq \delta$. 
Let $\mathbf X = \textup{vec}(\mathbf Y)$ where $\mathbf Y$ belongs to $(\Omega_Y\subset \mathbb{R}^{n\times n}, P_Y)$.
Then, if $\tilde\Omega_S$ denotes the set of all $n'\times n'$ substrings where $n'\leq n$, we have following bounds:
\begin{equation} \label{2Dbound1}
W(P_X, P_G) \leq
\left( n-n'+1 \right)^2\delta,
\end{equation}
and
\begin{equation}\label{2Dbound2}
W(P_X, P_G) \leq
\left( 1 + \frac{n-1}{n'} \right)^2\delta.
\end{equation}
\end{corollary}

 \begin{proof}
If $\tilde\Omega_S$ equals the set of all $n'\times n'$ substrings, then we have
\begin{equation}
|\tilde\Omega_S| = \left( n-n'+1 \right)^2 \; , \; \bm\lambda_{\min} = 1.
\end{equation}
Therefore, Theorem 1 gives
\begin{equation}
W(P_X, P_G) \leq \left( n-n'+1 \right)^2\delta
\end{equation}
which proves  (\ref{2Dbound1}). This result is also intuitive, because as the dimensions of substring ($n'\times n'$) tend to that of the sequence ($n\times n$), the bound on the distributional distance of sequences and substrings matches each other, i.e. $\delta$. See Fig. \ref{fig:corollary2}.

\begin{figure}[ht]
    \centering
    \includegraphics[width=\columnwidth]{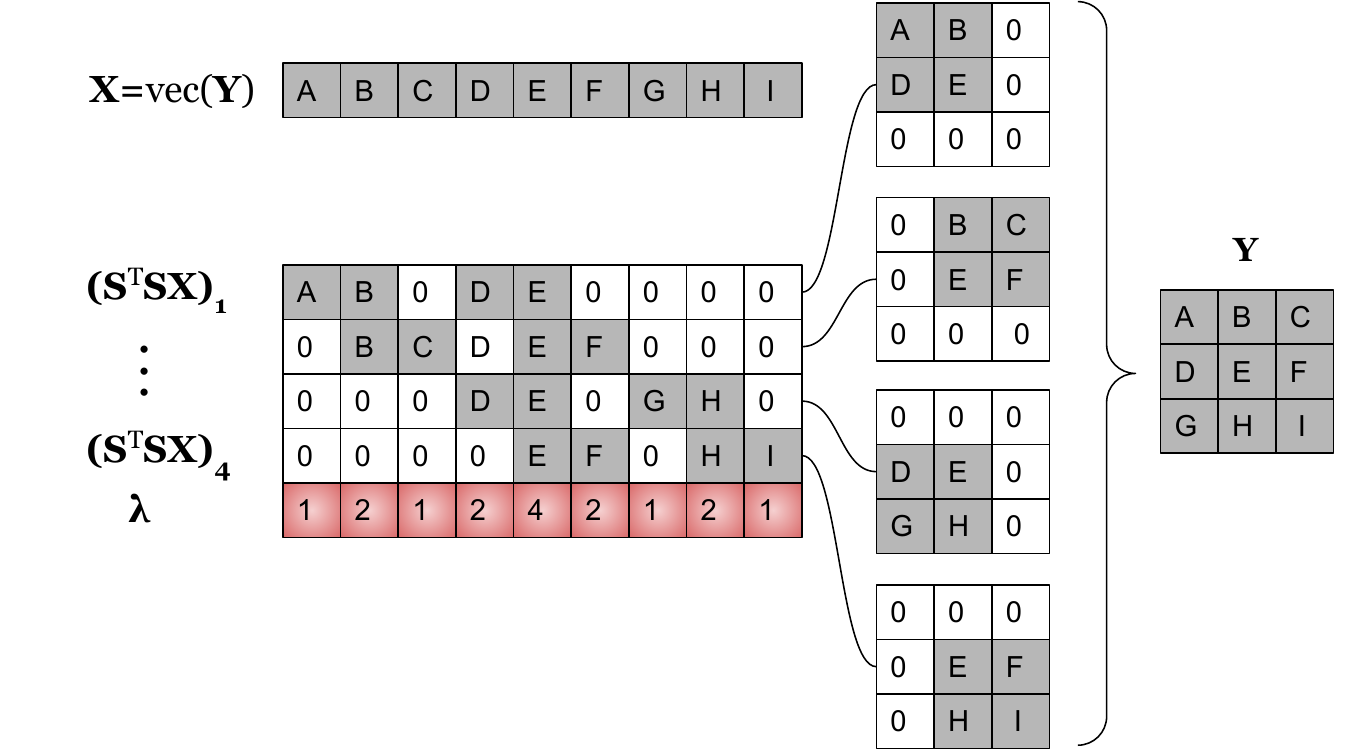}
    \caption{An $n\times n$ matrix $\mathbf Y$ with $n' \times n'$ substrings, vectorized as sequence $\mathbf X$ with projections as $\mathbf S^T \mathbf{SX}$, where $n=3$ and $n'=2$.}
    \label{fig:corollary2}
\end{figure}

Although the bound in  (\ref{2Dbound1}) is asymptotically intuitive (i.e. when $n' \rightarrow n$), we may obtain a tighter bound as follows.
It is evident that the previous bound suffers from $\bm\lambda_{\min} = 1$.
Alternatively, we may perform optimal transport on zero-padded data pairs, as zero elements do not contribute to the cost of transport between the distributions (Fig. \ref{fig:corollary2_2}).
On one hand, zero-padding a length of $n'-1$ per each side of $\mathbf Y$, gives $\min ( \bm\lambda) = n'^2$ for the set of non-padded elements.
On the other hand, $\bm\lambda$ can be assumed to have any (non-zero) value for the set of zero-padded elements as they do not contribute to the optimal transport. Therefore, for the padded $\mathbf{Y}$ we have:
\begin{equation}
|\tilde\Omega_S| = \left( n+n'-1 \right)^2 \; , \; \bm\lambda_{\min}= n'^2;
\end{equation}
which, using Theorem (1), gives:
\begin{equation}
\begin{split}
W(P_X, P_G) & \leq
\left( \left( n+n'-1 \right)^2 \mathbin{/} n'^2 \right)\delta \\ & = \left( 1+(n-1)\mathbin{/}n' \right)^2\delta.
\end{split}
\end{equation}
For $1<n'<n-1$,  (\ref{2Dbound2}) gives a tighter bound compared with  (\ref{2Dbound1}). While for $n'=1$ and $n'=n-1$ both give $n^2\delta$ and $4\delta$ as the upper bounds, respectively,  (\ref{2Dbound1}) is tighter only for the asymptotic case of $n'=n$.

\begin{figure}[ht]
    \centering
    \includegraphics[width=\columnwidth]{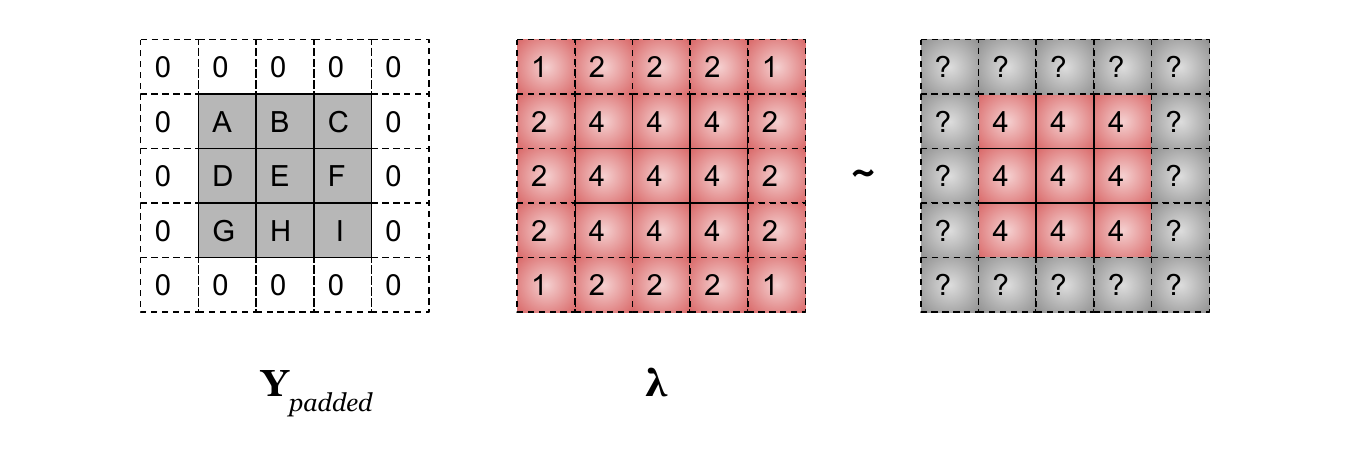}
    \caption{Zero-padded $\mathbf Y$ is $(n+2n'-2) \times (n+2n'-2)$, where $n=3$ and $n'=2$. The central $n\times n$ elements are repeated $n'\times n'$ times each. The repetition of other elements can be substituted arbitrarily (marked by ?), as corresponding elements in $\mathbf Y_{padded}$ are zero in both real and generated data and do not contribute to optimal transport.}
    \label{fig:corollary2_2}
\end{figure}

\end{proof}

\begin{remark}
\label{rem1}
In the preceding corollary, we presented results for square signal dimensions, i.e. $n\times n$, to maintain expositional clarity and enhance the elegance of the derivations. The bounds discussed can be readily extended to accommodate rectangular signal dimensions, such as $m\times n$ which involves analogous derivations. 
\end{remark}

\section{Experimental setup}
\subsection{Dataset}
RF sensing techniques for UAV detection and identification are practical for real-world use. These methods do not depend on UAVs' wireless protocols like Bluetooth, 4G, or WiFi, and can manage physical differences across UAV models. This study utilizes a UAV RF dataset \cite{al2019rf, allahham2019dronerf} which encompasses RF signals obtained from three distinct UAV variants across a range of flight scenarios and is readily available for public access\footnote{\url{https://github.com/Al-Sad/DroneRF}}. Specifically, the dataset supports three primary tasks: the detection of a UAV's presence (two classes), the detection of a UAV's presence coupled with the identification of its type (four classes), and the detection of a UAV's presence alongside the identification of its type and flight mode (ten classes). In this work, we skip the binary detection task as it may not be a challenge even for limited RF environments. For comprehensive details regarding the experimental setup—including the types of UAVs, the characteristics of the RF environments, and the data collection methodologies—readers are directed to the original publication. Addressing these aspects is beyond the scope of this work, which is focused on enhancing the performance of UAV identification using existing datasets.
\par
To facilitate the objectives of UAVs detection and identification, the authors employed Fourier transform to process the RF signals, thereby extracting their frequency components. RF signals are processed using Discrete Fourier Transform (DFT) for several key reasons. DFT reveals important frequency-domain characteristics often hidden in time-domain signals, reduces data dimensionality, and ensures uniform signal length, which is particularly beneficial when raw RF data may vary due to different UAV flight times. Furthermore, DFT facilitates the extraction of specific frequency features crucial for classification tasks and produces features that are generally more robust against common issues like Electromagnetic Interference (EMI) and Radio Frequency Interference (RFI) in UAV RF signals. Therefore, classifiers tend to perform better when trained on DFT-based features due to their inherent structure and discrimination power, leading to improved accuracy and generalization.
\par
The DFT-transformed RF signals in the referenced dataset consist of 2048 frequency features. For our application, we select the first 2025 features and organize them into a $45 \times 45$ matrix structure to facilitate one-shot generation. This specific choice serves to validate the theoretical bounds presented in Corollary \ref{cor2}. As noted in Remark \ref{rem1}, adopting a more general approach for rectangular dimensions, such as reshaping the signal into a $32 \times 64$ matrix, is also feasible. Additionally, omitting the last 23 frequency features from the total of 2048 has a negligible impact on the performance of the identification task. We also randomly selected 1,135 samples out of 22,700 (5\%)  from the dataset to simulate conditions of a low-data regime, ensuring data integrity and preserving the proportions of labels. Each sample in this dataset is associated with a label, which is a one-hot encoded vector sized either four or ten, depending on the classification task, with the first class designated for background noise in no-UAV scenarios.
\par
Initially, we apply a K-fold strategy (with $K=5$) to the original dataset, setting aside a consistent 20\% random sample for testing and verification in each fold, with proportional representation of each label. This method ensures that these test samples remain unused in the training phase, preserving their utility for evaluating model performance against adversarial samples. The training phase then exclusively employs the remaining 80\% of the dataset, totaling 908 samples in each fold which is called ``complete'' data throughout this work.
For a comprehensive analysis, we further segment the training portion of each fold into reduced datasets, representing five scenarios with sizes between $10\sim50\%$ of the training data. This segmentation allows us to demonstrate our model's effectiveness across a range of reduced training dataset sizes.
To ensure fair model performance comparisons, 5000 synthetic samples are created from subsampled sets for all generative models. This method facilitates meaningful evaluations of models on synthetic versus original data across varying scenarios.

\subsection{Generative models}
\subsubsection{Generative patch distribution matching (GPDM)}
The process of synthetic data generation, as demonstrated in Fig. \ref{fig:vrm}, utilizes one-shot generative models. Corollary \ref{cor2} establishes a theoretical bound on the distribution distance between the original dataset and the generated dataset. This ensures that while the dataset is expanded using one-shot generative techniques, the deviation from the original's distribution remains within a controlled limit. In implementing this method, we leverage architectures that are originally designed for single image generation, adapting them to treat the reshaped samples as $45\times 45$ images. This reshaping approach is substantiated by Corollary \ref{cor2} and is intended to align with the operational needs of the generative model. Following this step, we reverse the reshaping process for the subsequent phase, i.e. the classification task.

In this work, we use the algorithm in \cite{elnekave2022generating}, namely GPDM, which is based on minimizing the SWD between the patches of real and generated images.
The algorithm also utilizes a multi-scale hierarchical strategy to learn the global structure of the image in the early stages of training and its hyperparameter are detailed in Table \ref{table:gpdm}.

\begin{table}
\centering
\caption{Hyper-parameters for GPDM}
\label{table:gpdm}
\setlength{\tabcolsep}{3pt} % Adjust column spacing
\begin{tabular}{|l|l|}
\hline
Hyper-parameter & Value \\
\hline
Finest Scale & $45\times 45$ \\
Coarsest Scale & $15\times 15$ \\
Patch Size & $5\times 5$, $7\times 7$, $9\times 9$ \\
Decrease Rate of Scales & 0.8 \\
Number of Projections for SWD & 256 \\
Learning Rate & 0.01 \\
Iteration Steps & 500 \\
\hline
\end{tabular}
\end{table}

\subsubsection{Conditional deep generative models}

We compare conditional deep generative models trained on clean versus unclean data and augment training data using the mixup method to prevent overfitting, smooth labels, and enhance model generalization. Our examination extends to detailed architectures and strategies for Conditional Generative Adversarial Networks (CGAN) and Conditional Variational Autoencoders (CVAE), focusing on layer structures and hyperparameters.

Outlier sensitivity is a critical concern in the training of deep generative models, particularly in low-data regimes where the limited number of training samples can magnify the impact of anomalous data. Such anomalies can lead to mode collapse, a situation where the model fails to capture the diversity of the data distribution and instead focuses on a few modes, significantly degrading the model's performance. Removing outliers is thus essential to ensure that the training process accurately represents the underlying data distribution, promoting a more robust and generalizable model. Therefore, we compare the results of conditional deep generative models in two scenarios: using clean and unclean training data. The cleaning process employs the Median Absolute Deviation (MAD) method, using a threshold of 3 MADs to determine and remove outliers. For each class, outliers are identified by calculating the Euclidean norms (L2 norms) of the sample features. A sample is considered an outlier if the absolute deviation of its norm from the median norm exceeds the threshold of three times the MAD. This selective exclusion enhances the robustness and reliability of the models by ensuring that they are trained on the most representative data points.

In our approach to enhancing deep models' performance, we employ the mixup method ~\cite{zhang2017mixup} to expand our training dataset to 5000 samples. This process involves mixing pairs of data and label samples from the original dataset based on weights derived from a symmetric Beta distribution, specifically \text{Beta}($\alpha=0.2$, $\beta=0.2$). This augmentation strategy is critical before employing deep generative models, which require substantial amounts of data. By increasing our training dataset size through mixup, we address the high data demands of these models and promote label smoothing. This preparation helps mitigate overfitting and enhances the model’s generalization capabilities by exposing it to a more varied set of training examples, which is essential for effectively training deep generative models to learn robust features. Note that this augmentation strategy is specifically tailored to improve the training of deep generative models and is not directly applicable to classification tasks due to the generation of smooth labels that may obscure distinct class boundaries.

CGAN is a type of GAN that conditions its generation process on additional information like labels or data from other modalities. This conditioning leads to more controlled and targeted generation of data, allowing the network to produce specific types of outputs based on the given conditions~\cite{mirza2014conditional}. We enhance the implementation of CGAN by integrating Wasserstein GAN with Gradient Penalty (WGAN-GP), which improves training stability and mitigates mode collapse, thereby ensuring higher quality and diversity in the generated outputs ~\cite{gulrajani2017improved}. In our architecture of CGAN, the selection of depth and width for layers follows a pyramidical structure, with an increasing or decreasing format to control processing of information, where each layer refines the output of its predecessor. We also employ label injection across all linear layers, not just the initial ones; see Table \ref{table:cgan} where $C \in \{4,10\}$ denotes class size. This approach strengthens the labels' influence throughout the network. This integration of labels into the network's layers, as evidenced in literature, bolsters the conditioning effect, aligning outputs more closely with specified labels~\cite{brock2018large, devries2017modulating, miyato2018cgans}. Our hyperparameter configuration includes the use of the LeakyReLU activation function, set with a slope of 0.2 for negative input values, to prevent the ``dying ReLU" problem as well as to capture non-linearities in the data ~\cite{nair2010rectified,maas2013rectifier}. Batch size is determined as 64 and the learning rate for the Adam optimizer is set to 0.0005 for both discriminator and generator.

\begin{table}
\centering
\caption{Generator and Discriminator Architectures in CGAN}
\label{table:cgan}
\setlength{\tabcolsep}{2pt} % Adjust column spacing
\begin{tabular}{|c|c|c|c|c|c|c|c|} % Add vertical lines and adjust for 8 columns
\hline
\multicolumn{4}{|c|}{\textbf{Generator}} & \multicolumn{4}{c|}{\textbf{Discriminator}} \\ \hline
\textbf{Idx} & \textbf{Layer} & \textbf{Input} & \textbf{Output} & \textbf{Idx} & \textbf{Layer} & \textbf{Input} & \textbf{Output} \\ \hline
0 & Linear & 50+$C$ & 125 & 0 & Linear & 2025+$C$ & 500 \\
1 & LeakyReLU & - & - & 1 & LeakyReLU & - & - \\
2 & Linear & 125+$C$ & 250 & 2 & Linear & 500+$C$ & 250 \\
3 & LeakyReLU & - & - & 3 & LeakyReLU & - & - \\
4 & Linear & 250+$C$ & 500 & 4 & Linear & 250+$C$ & 125 \\
5 & LeakyReLU & - & - & 5 & LeakyReLU & - & - \\
6 & Linear & 500+$C$ & 2025 & 6 & Linear & 125+$C$ & 1 \\
7 & Tanh & - & - & - & - & - & - \\ \hline
\end{tabular}
\end{table}

CVAE, a variant of VAE, distinctively incorporates conditional data such as labels or contextual inputs to steer its generative mechanism. This incorporation of conditional information is central to CVAE's functionality, enabling it to generate data that is both precise and contextually relevant, allowing the network to create outputs finely tuned to the specified conditions. Similar to CGAN, our CVAE architecture embodies a tiered layer structure, mirroring a pyramid-like progression in the depth and width of layers with labels incorporated in all linear layers; see Table \ref{table:cvae}. CVAE utilizes a dual-component loss function: a reconstruction loss, ensuring the fidelity of the generated data to the original input, and a Kullback-Leibler (KL) divergence loss, which maintains the efficiency and effectiveness of the latent space representation ~\cite{sohn2015learning,kingma2013auto}.
To dynamically balance the contribution of the KL divergence loss across training epochs, the CVAE employs a cyclical annealing schedule, with the KL weight peaking at its maximum value of 1.0 every 10 epochs. This sinusoidal modulation facilitates a smoother and more controlled integration of the KL divergence into the overall loss, enhancing the stability and quality of the latent space learning process ~\cite{fu2019cyclical}. The configurations for our model include using a LeakyReLU activation function with a slope of 0.2, a batch size of 64, and an Adam optimizer with a learning rate of 0.0001.

\begin{table}
\centering
\caption{Encoder and Decoder Architectures in CVAE}
\label{table:cvae}
\setlength{\tabcolsep}{2pt} % Adjust column spacing
\begin{tabular}{|c|c|c|c|c|c|c|c|} % Add vertical lines and adjust for 8 columns
\hline
\multicolumn{4}{|c|}{\textbf{Encoder}} & \multicolumn{4}{c|}{\textbf{Decoder}} \\ \hline
\textbf{Idx} & \textbf{Layer} & \textbf{Input} & \textbf{Output} & \textbf{Idx} & \textbf{Layer} & \textbf{Input} & \textbf{Output} \\ \hline
0 & Linear & 2025+$C$ & 500 & 0 & Linear & 50+$C$ & 125 \\
1 & LeakyReLU & - & - & 2 & LeakyReLU & - & - \\
2 & Linear & 500+$C$ & 250 & 4 & Linear & 125+$C$ & 250 \\
3 & LeakyReLU & - & - & 6 & LeakyReLU & - & - \\
4 & Linear & 250+$C$ & 125 & 8 & Linear & 250+$C$ & 500 \\
5 & LeakyReLU & - & - & 10 & LeakyReLU & - & - \\
6 & Linear & 125+$C$ & 2$\times$50 & 12 & Linear & 500+$C$ & 2025 \\
- & - & - & - & 13 & Tanh & - & - \\ \hline
\end{tabular}
\end{table}

\subsection{Classifier}

We opt for a pipeline that integrates the Synthetic Minority Over-sampling Technique (SMOTE) ~\cite{chawla2002smote, imblearn2023} with XGBoost's Classifier ~\cite{chen2016xgboost,xgboost2023}. This strategic choice targets class imbalance, a pervasive challenge in machine learning, using SMOTE to bolster the representation of minority classes effectively. Coupled with XGBoost, renowned for its efficiency, scalability, and adaptability across various applications, we employ an ensemble method that significantly enhances accuracy and generalization. This pipeline aligns seamlessly with our overarching research priorities, allowing to channel our efforts and resources more effectively into the core aspect of our work - synthetic data generation - while still ensuring reliable classification results.

In our setup, 20\% of the training data is reserved for five random validation sets, over which results are averaged to enhance robustness and reliability for every case. We incorporate a patience of 10 evaluations during validation to determine early stopping, further refining our model's performance. Our model is configured with the ``multi:softprob" objective, suitable for multiclass problems as it outputs a probability for each class. It evaluates using the ``mlogloss" and ``merror" metrics; ``mlogloss" measures the accuracy of the probability outputs while ``merror" tracks the misclassification rate, together providing a comprehensive view of model performance across multiple classes.

\subsection{Evaluation metrics}
The evaluation is conducted on ten tasks: 4-classes and 10-classes UAV (and flight mode) identification, each having five cases of generation on 10, 20, 30, 40, and 50\% of the complete data.
For each case, we use a randomly sampled 5-fold cross-validation \cite{zeng2000distribution} setting and report the average scores over all of the folds.
For the evaluation of multilabel classification in every scenario, the accuracy, precision, recall, and F1-score are computed for each label and subsequently averaged based on the support of each label (the number of true instances for each label).

\section{Results and discussion}
\subsection{Validation of bounded divergence}

In Section \ref{similarities}, we established an upper bound for the distributional distance between synthetic and real data, focusing on the distribution of patches within samples. Specifically, the bound presented in Corollary \ref{cor2} ensures that two-dimensional arrays, such as images, generated by one-shot generative models like GPDM, do not exhibit a distributional divergence from real arrays beyond a predefined threshold. In this subsection, we demonstrate that such threshold holds in practice.

\begin{figure}
    \centering
    \includegraphics[width=1\columnwidth]{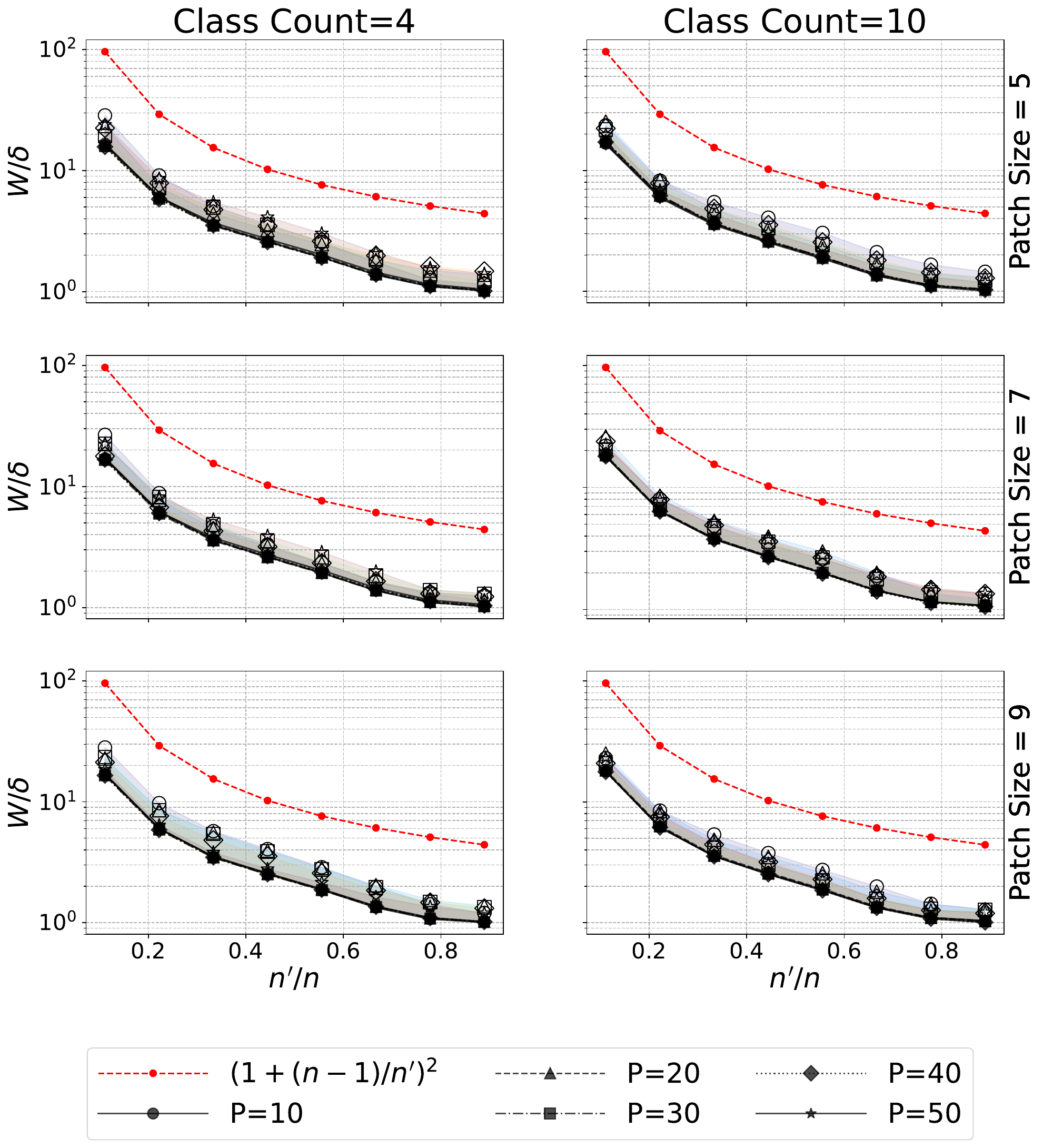}
    \caption{Mean, as well as mean plus ten times the standard deviation of $W/\delta$ (shaded area), calculated over five folds for different scenarios to verify its compliance with the theoretical upper limit in Eq. (\ref{2Dbound2}), where P denote the percentage of utilized data.}
    \label{fig:bounds}
\end{figure}

Let $W$ represent the distributional distance between synthetic and real datasets, and $\delta$ denote the expected distributional distance between their respective patches. We demonstrate the relationship of $W/\delta$ to its upper bound, as detailed in Eq. (\ref{2Dbound2}). Fig. \ref{fig:bounds} confirms that  Eq. (\ref{2Dbound2}) is applicable across various scenarios, encompassing different class counts, patch sizes used in generation, proportions of real data utilized, and the patch size $(n')$ employed for calculating $\delta$. The figure presents the mean of $W/\delta$, calculated over five folds, and also shows this mean value plus ten times the standard deviation of $W/\delta$.

Despite the consistent applicability of the bound across all scenarios, it may not always appear tight. In this work, the observed phenomenon can primarily be attributed to the multiscale approach of GPDM. This approach initially concentrates on capturing the global structure of samples at a lower scale. This process effectively reduces $W$ to a greater extent. Also, $\delta$ is subject to overestimation using limited patches within each sample, because the Wasserstein distance is sensitive to the tails of distributions. Limited samples may not adequately represent these tails, leading to an overestimation of the actual distributional distance. Additionally, in high-dimensional spaces, the curse of dimensionality can inflate perceived differences, further contributing to overestimation. Sampling variability, especially when capturing more extreme values or outliers, can also skew the perceived distribution, resulting in an overestimated Wasserstein distance between the respective patches $(\delta)$.

Therefore, we have successfully established and validated an upper bound for the distributional distance between synthetic and real data using one-shot generative models, specifically focusing on GPDM. Our investigations confirm the practical applicability of the established bound across a range of scenarios, as demonstrated in the consistency of $W/\delta$'s adherence to its theoretical upper limit. However, it is important to recognize that certain factors, such as the multiscale nature of GPDM and the limited patch counts within each sample, can lead to the underestimation of $W/\delta$'s in empirical circumstance.

\subsection{Classification performance}

We can evaluate the identification of UAV types and flight modes by employing synthetic RF transforms across various scenarios. These scenarios are distinguished by the proportion of data used in the synthesis process and the generation methodology. Specifically, we assess classification performance by comparing a complete dataset and its subsets sampled at different proportions with their corresponding synthetically generated counterparts, utilizing a range of models.

\begin{figure}
    \centering
    \includegraphics[width=0.8\columnwidth]{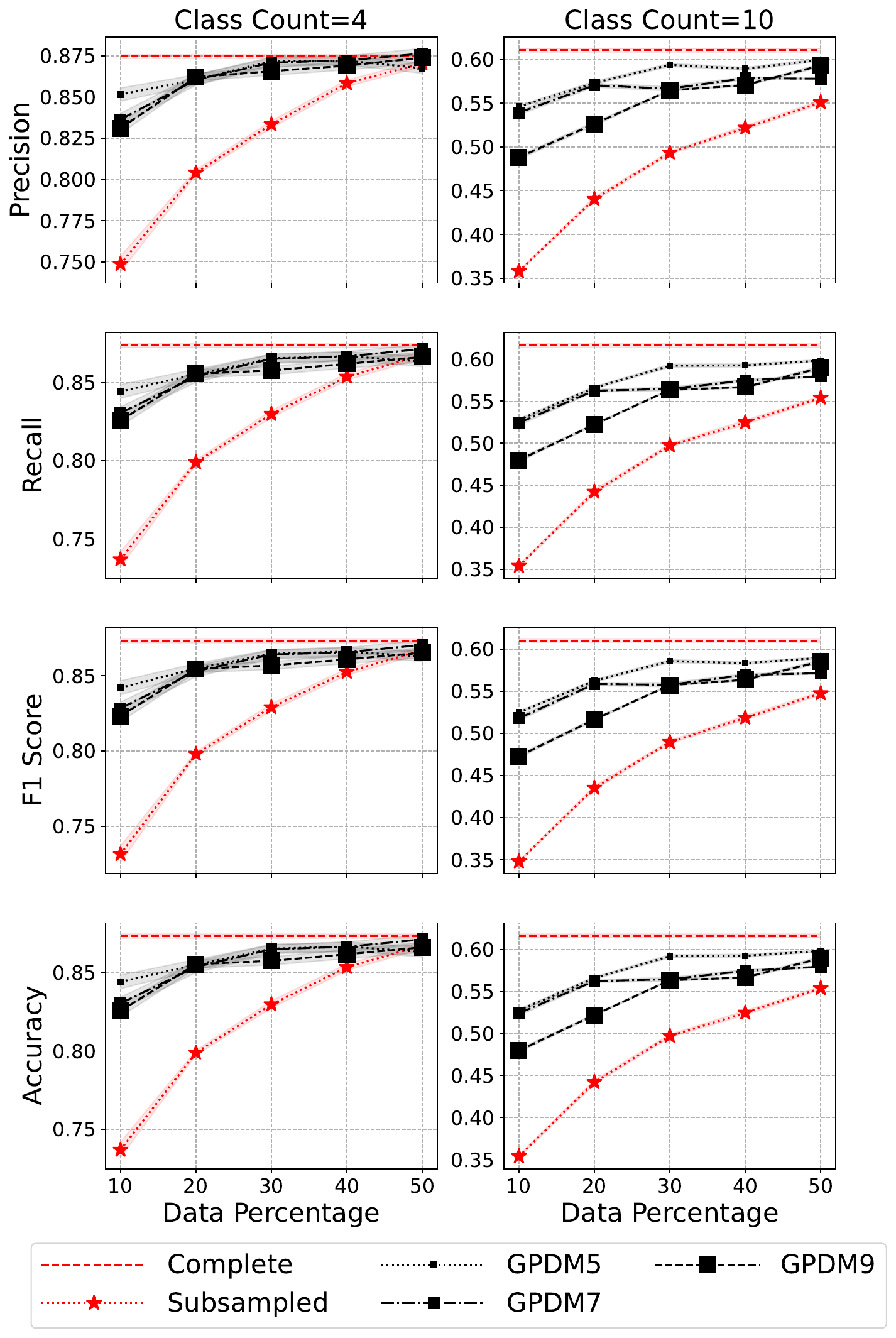}
    \caption{The performance of UAV identification using synthetic transformed RF signals generated by GPDM.}
    \label{fig:perf_gpdm}
\end{figure}

\begin{figure}
    \centering
\includegraphics[width=0.8\columnwidth]{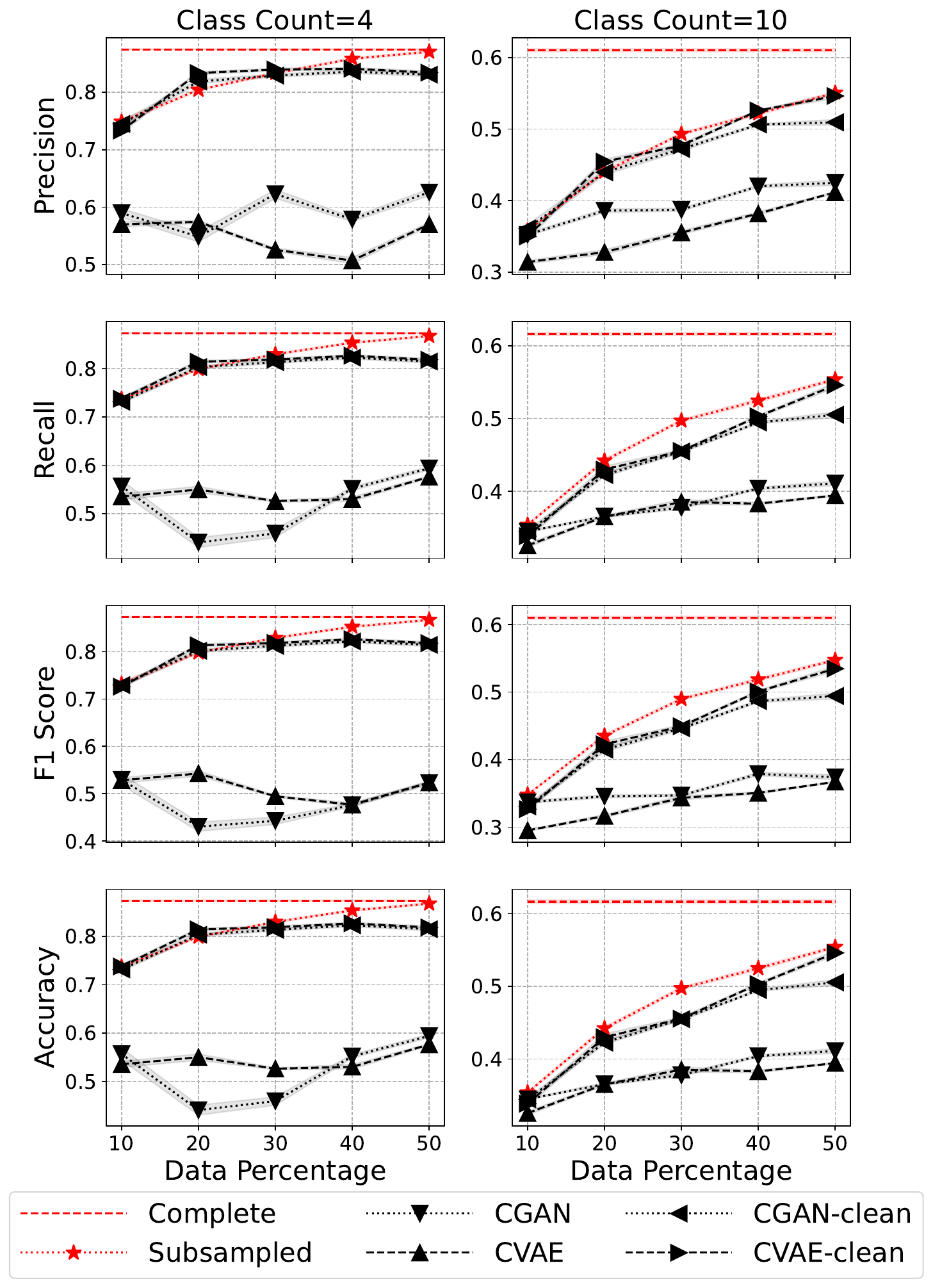}
    \caption{The performance of UAV identification using synthetic transformed RF signals generated by CGAN and CVAE.}
    \label{fig:perf_deep}
\end{figure}

The classification performance metrics are presented in Figs. \ref{fig:perf_gpdm} and \ref{fig:perf_deep}, which both incorporate results from complete and subsampled datasets for comparison. In this context, the complete datasets serve as our ground truth, while the subsampled datasets function as baselines. The metrics for the complete data remain constant since data proportion is not a factor. On the other hand, the metrics for the subsampled data improve with the use of more data.

Fig. \ref{fig:perf_gpdm} illustrates the performance metrics of classification using synthetic RF transformations generated by the one-shot GPDM approach. Generally, the performance metrics of the synthetic data fall between those of the baseline and the ground truth, demonstrating the effectiveness of the one-shot generative approach for data synthesis. The impact of patch size on various metrics is significant, especially in the 10-class scenario where utilizing a patch size of 5 is more successful than larger patch sizes in expanding the support of class-wise distributions. The benefit of choosing a smaller patch size becomes less pronounced as the proportion of data accessed increases (approaching 50\%), where further extrapolation of distributional support no longer offers additional benefits.

Fig. \ref{fig:perf_deep} illustrates the performance metrics of classifications using synthetic RF transformations generated by conditional deep learning models such as CGAN and CVAE. Contrary to the GPDM approach, CGAN and CVAE underperform, often not even reaching the baseline metrics, primarily due to their inefficiency in scenarios with limited data. In such low-data regimes, the outcome where deep learning models tend to memorize small training samples represents the best-case scenario. This already precarious situation is further aggravated if the training data is not meticulously clean. The presence of noise and outliers in the data poses significant challenges, making it a struggle for these models to even achieve basic memorization of the training samples, let alone generalize effectively.

One-shot generative data augmentation, in contrast to CGAN and CVAE, excels particularly in environments with outliers. Its robustness arises from its capacity for internal learning, where the model extracts and utilizes the intrinsic characteristics of a single example to generate a broad range of synthetic samples. This approach focuses on the core attributes of the sample, effectively minimizing the influence of outliers. By capturing and extrapolating the essential features, the one-shot method not only improves the diversity and representativeness of the synthetic data but also significantly enhances its utility for training models in scenarios plagued by data anomalies.

The comparative analysis, depicted in Fig. \ref{fig:perf_gpdm} and Fig. \ref{fig:perf_deep}, reveals significant insights into the performance of the one-shot generative approach using GPDM and conditional deep generators such as CGAN and CVAE. While the one-shot GPDM method consistently demonstrates superior handling and generation of synthetic data in the presence of outliers and limited data scenarios, the performance of CGAN and CVAE lags, especially when data cleanliness and abundance are issues. This study underscores the critical importance of choosing appropriate data synthesis techniques based on the specific challenges and data environment encountered. It suggests that one-shot generative methods, with their robust internal learning mechanisms, are not only more resilient but also more adaptable to diverse and challenging data landscapes, making them a preferable choice in situations where data quality cannot be guaranteed.
\section{Limitations and Future Works}
The primary focus of our research was to address data scarcity in RF-based UAV identification through one-shot generative data augmentation, but we did not analyze how individual environmental factors affect performance. Future work could involve a detailed analysis of the model’s performance against datasets that incorporate factors such as multipath propagation, non-line-of-sight conditions, heavy urban interference, or severe weather individually, providing a more comprehensive evaluation of the method’s robustness. Additionally, incorporating a physics-based approach could enhance the generation of more realistic samples, allowing for an insightful understanding of the RF environment's impact on signal integrity.

While one-shot generative data augmentation shows promise in low-data regimes, its computational intensity could pose challenges in specific operational contexts. This issue stems from the inherent complexity of one-shot generative models, suggesting that future research should focus on optimizing these algorithms and exploring high-performance computing techniques. Such advancements could enhance the model’s applicability to real-time identification tasks, also facilitating more extensive studies using datasets that capture varied real-world conditions, thereby further validating its robustness and practical applicability.

Lastly, the ethical considerations in deploying one-shot generative data augmentation, particularly in UAV identification, also warrant careful attention. Ensuring that synthetic data does not inadvertently introduce bias or inaccuracies that could compromise UAV detection or misclassify benign entities as threats is paramount. Furthermore, as these models may be employed in security-sensitive areas, the implications of errors—whether false positives or negatives—could have serious repercussions. Therefore, it is essential to establish rigorous validation frameworks and transparency in model training processes to maintain trust and reliability in the use of synthetic data in UAV identification systems.

\section{Conclusion}
This study demonstrates that one-shot generative models, such as GPDM, significantly improve UAV classification metrics, especially in limited data scenarios where traditional deep generative models struggle.
We rigorously evaluated the application of one-shot generative approaches in synthesizing RF signal transforms, establishing and validating an upper bound for the distributional distance between synthetic and real data. Our results consistently align with theoretical predictions across various scenarios, confirming the generation robustness in distribution matching. This validation ensures the synthetic data's fidelity for reliable UAV identification. Furthermore, one-shot generative augmentation outperforms CGAN and CVAE in classifying UAVs and their flight mode with limited data and high signal complexity. Overall, our research offers a rigorous and effective approach to improve UAV identification in limited RF environments.

\section*{Acknowledgment}
This work utilizes resources supported by the National Science Foundation’s (NSF) Major Research Instrumentation program, grant \#1725729, as well as the University of Illinois at Urbana-Champaign ~\cite{kindratenko2020hal}. The first author also appreciates NSF's NCSA Internship Program for Cyberinfrastructure Professionals, grant \#1730519 ~\cite{lapine2020ncsa}.

%\end{linenumbers}
\newpage

% Insert the bibliography
\bibliographystyle{elsarticle-num} % Elsevier style
\bibliography{references} % Your .bib file

\end{document}